\documentclass{article} 
\usepackage{iclr2023_conference,times}

\usepackage{amsmath,amsfonts,bm}

\def\eqref#1{equation~\ref{#1}}

\def\1{\bm{1}}

\DeclareMathAlphabet{\mathsfit}{\encodingdefault}{\sfdefault}{m}{sl}
\SetMathAlphabet{\mathsfit}{bold}{\encodingdefault}{\sfdefault}{bx}{n}

\usepackage{hyperref}
\usepackage{url}
\usepackage{subfigure}
\usepackage{graphicx}
\usepackage{algorithm}
\usepackage{algpseudocode}
\usepackage{wrapfig}
\usepackage{tablefootnote}
\usepackage[flushleft]{threeparttable}
\usepackage{multirow}
\usepackage{nicefrac}
\usepackage{enumitem}
\usepackage{booktabs}     
\usepackage{makecell}
\usepackage{bbm}

\title{DCE: Offline Reinforcement Learning With Double Conservative Estimates}

\author{Chen Zhao$^{1}$ \quad Kaixing Huang$^{1}$\quad Chun Yuan\textsuperscript{$\dagger$}$^{1}$\\
$^{1}$Tsinghua Shenzhen International Graduate School\\
{\tt\small \{zhao-c21@mails, yuanc@sz\}.tsinghua.edu.cn \quad}\\
}

\iclrfinalcopy 
\begin{document}

\maketitle
\renewcommand{\thefootnote}{\fnsymbol{footnote}}
\footnotetext[2]{Corresponding author}

\begin{abstract}
Offline Reinforcement Learning has attracted much interest in solving the application challenge for
traditional reinforcement learning. Offline reinforcement learning uses previously-collected datasets to train agents without any interaction. For addressing the overestimation of OOD (out-of-distribution) actions, conservative estimates give a low value for all inputs. Previous conservative estimation methods are usually difficult to avoid the impact of OOD actions on Q-value estimates. In addition, these algorithms usually need to lose some computational efficiency to achieve the purpose of conservative estimation. In this paper, we propose a simple conservative estimation method, double conservative estimates (DCE), which use two conservative estimation method to constraint policy. Our algorithm introduces V-function to avoid the error of in-distribution action while implicit achieving conservative estimation. In addition, our algorithm uses a controllable penalty term changing the degree of conservatism in training. We theoretically show how this method influences the estimation of OOD actions and in-distribution actions. Our experiment separately shows that two conservative estimation methods impact the estimation of all state-action. DCE demonstrates the state-of-the-art performance on D4RL. 
\end{abstract}

\section{Introduction}

Reinforcement Learning has made significant progress in recent years. However, in practice, the application of reinforcement learning is mainly in games that can easily interact with the environment. Although off-policy reinforcement learning uses previously-collected datasets when it trains agents, in practice, off-policy does not perform well without any interaction. In order to apply reinforcement learning in those environments which are hard to interact with (e.g. autonomous driving,  medical care), offline reinforcement learning has attracted some attention. Offline reinforcement learning aims to train an agent by previous-collected, static datasets without further interaction and wishs a good performance of this agent which can be competitive with those online algorithm's agent.

The main challenge for offline reinforcement learning is that critics tend to overestimate the value of out-of-distribution state-action, which is introduced by the distribution shift. Another problem of offline reinforcement learning is that it cannot improve exploration due to the absence of environmental interaction. Unfortunately,  this problem cannot be solved because of the nature of the model-free offline reinforcement learning algorithm (\cite{levine2020offline}). Therefore, the offline reinforcement learning algorithm mainly addresses the overestimation caused by distribution shifts. There are many methods to solve the overestimation caused by distribution shifts. We can simple divide offline RL algorithm into the following categories:Policy constraint (e.g. BCQ \cite{fujimoto2019off},BEAR \cite{kumar2019stabilizing}), Conservative estimation (e.g. CQL \cite{kumar2020conservative}), Uncertainty estimation (e.g. PBRL \cite{bai2022pessimistic}), Using DICE (e.g. AlgaeDICE \cite{nachum2019algaedice}), Using V-function (e.g. IQL \cite{kostrikov2021offline}), Other (e.g. DR3 \cite{kumar2021dr3}),etc.

In those algorithms, IQL solve a problem which is caused by Bellman equation: $\mathcal{B}^{*} Q(\mathbf{s},  \mathbf{a})=r(\mathbf{s},  \mathbf{a})+\gamma \mathbb{E}_{\mathbf{s}^{\prime} \sim P\left(\mathbf{s}^{\prime} \mid \mathbf{s},  \mathbf{a}\right)}\left[ Q\left(\mathbf{s}^{\prime},  \mathbf{a}^{\prime}\right)\right]$ ($a^\prime$ comes from the dataset). IQL learn a V-value of state by expectile regression to control the estimation of state-action. By using advantage-weighted regression, the policy of IQL can cover the optimal in-distribution action. However, this means that IQL is too 'conservation' to discover the optimal action, which may be near the in-distribution action. On the other hand, previous conservative estimation methods need to lose some computational efficiency to achieve the purpose of conservative estimation.

Our work proposes a method to explore the OOD action to get a better policy appropriately while keeping the Q-values of in-distribution actions relatively accurate. We found that the V-function has a conservative estimation effect, but it is not enough (see Appendix E). Therefore, we need additional conservative estimation to constrain the exploration of OOD action. Our algorithm uses a simple conservative estimation algorithm without requiring pre-training like Fisher-BRC. We use V-function and an additional penalty term to achieve conservative estimation. In addition, our algorithm uses a controllable parameter to change the degree of conservation in training. Our algorithm aims to achieve limited exploration of OOD action by conservative estimation while having little influence on the state-action of the dataset. Our algorithm does not require additional pre-training and maintains higher computational efficiency. In addition, our algorithm was only modified a little for the standard Actor-Critic algorithm. Our main contributions are  as follows:
\begin{itemize}
    \item We propose that using V-function and an additional term limit the exploration of policy to close to the distribution of critics and alleviate the impact of OOD action on in-distribution action. In addition, our experiment shows the influence of V-function.
    \item We propose a flexible conservative estimation algorithm in training, which does not require additional training. Specifically, our algorithm only requires adding one penalty term to the critic's loss objective to implement the above goal.
    \item We theoretically demonstrate how this method influences the estimation of OOD actions and in-distribution actions. In addition, we did experiments to show the effect of our algorithm in practice.
    \item We experimentally demonstrate that our algorithm can achieve state-of-the-art performance on D4RL.
\end{itemize}

\section{Related Work}
\label{gen_inst}
In recent years, many offline reinforcement learning algorithms have been studied to apply reinforcement learning to more practical fields other than games. We classify those by the method of their algorithm as follows:
\begin{itemize}
    \item \textbf{Policy constraint:} One method to achieve offline RL is policy constraint, which contains implicit and explicit constraints. Explicit constraint direct constrain agent's policy into behavior policy which is using to collect datasets, including: BCQ (\cite{fujimoto2019off}), DAC-MDP (\cite{shrestha2020deepaveragers}), BRAC (\cite{YifanWu2019BehaviorRO}), EMaQ (\cite{SeyedKamyarSeyedGhasemipour2020EMaQEQ}), etc. Implicit constraint use implicit method, like f-divergence , to constrain policy close to behavior policy including: BEAR (\cite{kumar2019stabilizing}), ABM (\cite{siegel2020keep}), AWAC (\cite{AshvinNair2022AWACAO}), AWR (\cite{XueBinPeng2021AdvantageWeightedRS}), CRR (\cite{wang2020critic}), etc.
    \item \textbf{Conservative estimation:} This method avoids policy selecting OOD action by giving a lower value of OOD state-action because overestimation and irremediable value cause policy extrapolation errors. It include: CQL (\cite{kumar2020conservative}) Fisher-BRC (\cite{fisher2021}), etc.
    \item \textbf{Uncertainty estimation:} This is another method to avoid overestimation by giving the uncertainty of estimation of critic or model. The algorithm gives an uncertainty of estimation, and if it exceeds the threshold, the algorithm will take specific measures. it include: MOReL (\cite{kidambi2020morel}), PBRL (\cite{bai2022pessimistic}), UWAC (\cite{wu2021uncertainty}), etc.
    \item \textbf{Using V-function:} Another method is using the V-function of state to avoid overestimating action because V-value does not need action. It also constrains policy, but we separate it from policy constraint because it uses a different Bellman equation. These methods usually use advantage-weighted behavioural cloning to prevent extrapolation. It include: IQL (\cite{kostrikov2021offline}), VEM (\cite{ma2021offline}), etc.
    \item \textbf{Using DICE:} The above algorithm does not directly solve the distribution shift problem. DICE solve this problem by calculating discounted stationary distribution rations. It include: AlgaeDICE (\cite{nachum2019algaedice}), OPtiDICE (\cite{lee2021optidice}), RepB-SDE (\cite{lee2020representation}).
    \item \textbf{Other:} there are some algorithm which don't fall into above category: DR3 (\cite{kumar2021dr3}), OPAL (\cite{ajay2020opal}), O-RAAC (\cite{urpi2021risk}), etc.
\end{itemize}
Our work focuses on the conservative estimation algorithm and v-function to get good performance while maintaining the algorithm's efficiency and concise.

In our work, we try to solve the valuation error of conservative estimation introduced by the Bellman equation. Suppose the constraint of policy use advantage-weighted behavioural cloning; the agent may miss optimal action near the existing policy distribution. However, using V-value without any policy constraint will get a weak performance because of OOD action. Therefore, we need another method to constrain the policy, which avoids the valuation error of conservative estimation and gives a conservative estimation for OOD action. In addition, we control the degree of conservation by updating the parameter in training. Our algorithm includes the quantile regression (IQL, VEM) and conservative estimation. The effect of quantile regression has been demonstrated by \cite{kostrikov2021offline} and \cite{ma2021offline}. We simply introduce the quantile regression content and show its influence on our algorithm. Section 4 mainly shows our penalty term and analyses what it does for our algorithm.

\section{Preliminaries}
\label{headings}

The goal of Reinforcement Learning is to train a policy to maximize the expected cumulative reward, which is usually set by experts. Many RL algorithms are considered in Markov decision process (MDP), which contains a tuple $(\mathcal{S},  \mathcal{A},  T,  r,  \gamma)$.$\mathcal{S},  \mathcal{A}$ represent state spaces of environment and action spaces of policy, $T\left(\mathbf{s}^{\prime} \mid \mathbf{s},  \mathbf{a}\right)$ is environment dynamic and tell us the transition probability of state s' when policy use action a in state s. $r(\mathbf{s},  \mathbf{a})$ represents reward function, and $\gamma \in(0, 1)$ is discount factor. $\pi_{\beta}(\mathbf{a} \mid \mathbf{s})$ represents behavior policy which use to collect dataset D, $d^{\pi_{\beta}}(\mathbf{s})$ is discount marginal state-distribution of $\pi_{\beta}(\mathbf{a} \mid \mathbf{s})$. The goal of the policy is to maximize returns: 
\begin{equation}
\pi^{*}=\underset{\pi}{\arg \max } \mathbb{E}_{\pi}\left[\sum_{t=0}^{\infty} \gamma^{t} r\left(s_{t},  a_{t}\right) \mid s_{0} \sim p_{0}(\cdot),  a_{t} \sim \pi\left(\cdot \mid s_{t}\right),  s_{t+1} \sim p\left(\cdot \mid s_{t},  a_{t}\right)\right]
\end{equation}
The off-policy algorithm achieves the above objectives by utilising the state-action value function  (Q-function) Q(s, a).

\textbf{Offline reinforcement learning}. In contrast to the online RL algorithm, offline RL uses previously static collected data training policy without further environment interaction. Recently many model-free offline RL algorithms train critics by minimizing temporal difference error and its loss: 
\begin{equation}
L_{T D}(\theta)=\mathbb{E}_{\left(s,  a,  s^{\prime}\right) \sim \mathcal{D}}\left[\left(r(s,  a)+\gamma \max _{a^{\prime}} Q_{\hat{\theta}}\left(s^{\prime},  a^{\prime}\right)-Q_{\theta}(s,  a)\right)^{2}\right]
\end{equation}
where D is the dataset, $a^{\prime}$ represents the action which is sample from policy, $Q_\theta(s, a)$ is a parameterized Q-function, $Q_{\hat{\theta}}(s^{\prime}, a^{\prime})$ is target network (usually use soft parameters update), $s^{\prime}$ is next state and $a^{\prime}$ represents action which sample from $\pi$. As we can see, if $a^{\prime}$ is an OOD action, Q-function does not estimate it correctly, and TD loss will exacerbate this error. Overestimation caused by accumulative error will lead to policy extrapolation, making it difficult for performance to converge to a good level. So, many offline RL algorithms add a penalty term to solve the problem that Q-function overestimates OOD state-action.

\section{Double Conservative Estimates}
In this section, we will show our work, which aim is to try to get a relatively accurate in-dataset action estimation and reduce the overestimation value of OOD state-action in training. Therefore, our algorithm must penalise the overestimation of OOD action while avoiding the influence of OOD action on the in-distribution action.

First, for the traditional conservative estimation offline RL algorithm, the critic uses the following target loss function: 
\begin{equation}
L_{T D}(\theta)=\mathbb{E}_{\left(s,  a,  s^{\prime}\right) \sim \mathcal{D}}\left[\left(r(s,  a)+\gamma \max _{a^{\prime}} Q_{\hat{\theta}}\left(s^{\prime},  a^{\prime}\right)-Q_{\theta}(s,  a)\right)^{2}\right] + penalty term \label{eq:Q_L}
\end{equation}
$a^\prime$ is a sample from the policy. If this action is an OOD action, Q-function will give a wrong Q-value which is difficult to correct because of the nature of offline RL. In addition, the penalty term must be carefully designed to avoid a poor performance introduced by an inappropriate conservative estimation penalty term design.

Unlike the traditional reinforcement learning algorithm that uses conservative estimation, our algorithm can achieve the above goal without introducing additional agent training. Specific, our algorithm uses the V-function mentioned above to train Q-function and add a penalty term to control the degree of conservatism estimation. Our algorithm reduces the Q-value of the OOD state-action and ensures that the values of the in-distribution state-action will not introduce additional errors.

\subsection{Learning Q-function with V-funcion}

Our work use SARSA-style objective to estimate in-dataset state-action which has been considered in prior Offline Reinforcement Learning (\cite{kostrikov2021offline}, \cite{ma2021offline}, \cite{brandfonbrener2021offline}, \cite{gulcehre2021regularized}) to achieve the first goal. We train V-function use the following loss: 
\begin{equation}
L_{V}(\psi)=\mathbb{E}_{(s,  a) \sim \mathcal{D}}\left[L_{2}^{\tau}\left(Q_{\hat{\theta}}(s,  a)-V_{\psi}(s)\right)\right] \label{eq:V_L}
\end{equation}
where $L_{2}^{\tau}(u)=|\tau-1(u<0)| u^{2}$. As a result, this equation balances the weights of different values of Q concerning values of V. The corresponding loss of Q-function is as follow: 
\begin{equation}
L_{Q}(\theta)=\mathbb{E}_{(s,  a,  s^{\prime}) \sim \mathcal{D}}[(r(s,  a)+\gamma V_{\psi}(s^{\prime})\\-Q_{\theta}(s,  a))^{2}] \label{eq:QL_tau}
\end{equation}
The advantage of the above equation is that this equation can eliminate the relationship between the action of the current state and the action of the next state, which can avoid the problem of the bellman equation. In addition, using V-function can appropriately alleviate the problem of overestimation, which we will show experimentally. Furthermore, we see this as a conservation method.

The critic with V-function can cover the in-dataset optimal policy distribution. However, only having an accurate in-dataset action estimation cannot effective relieve overestimation of OOD action, so we need to reduce the estimation of OOD action while sustaining relatively accurate in-dataset action estimation. 

\subsection{OOD Penalty Term}
The idea of our algorithm is that for the Q-value of OOD state-action, we do not need to get its exact conservative value but only need its value lower than the value of optimal action strategies to exist in the dataset. For values of OOD state-action that are lower than the values of in-distribution action, we do not care. For values of OOD state-action that may be higher, caused by overestimation, than the values of in-dataset action, we hope to bring it down to a lower level. At the same time, the agent can also select better action.

We propose a simple penalty term to achieve the above aim. Our work uses Q-value off OOD action (sampled by current policy) as a penalty term to reduce overestimation. At last,  Q-function's objective loss is as follows: 
\begin{equation}
L_{Q}(\theta)=\mathbb{E}_{\left(s,  a,  s^{\prime}\right) \sim \mathcal{D}}\left[\left(r(s,  a)+\gamma V_{\psi}\left(s^{\prime}\right)-Q_{\theta}(s,  a)\right)^{2}\right]+\beta\times\mathbb{E}_{s \sim \mathcal{D}, a^{\prime} \sim \pi_\theta}[Q_\theta(s, a^{\prime})] \label{eq:our}
\end{equation}
In this objective, $Q_{\hat{\theta}}(s, a^{\prime})$ represents Q-value of state-action $(s, a^{\prime})$ which $a^\prime$ sample from current policy. In this penalty term, our work uses the current policy as an OOD action sampler rather than the randomly initialized policy because low-quality action will push Q-function in the wrong direction. In the experiment, we use the current policy to generate the action of the current state, taking its Q value as the penalty term. We will analyze the influence of this penalty term and give a lower bound theory.

\subsection{Analysis}
To facilitate analysis, we set the parameter $\beta$ to a fixed value, which follows the train regularly. In practice, however, a variable $\beta$ may have a better performance in some datasets because of the impact of conservative estimation on policy. Furthermore, for variable $\beta$, the result of Q-value and V-value are the same as those of constant $\beta$.

First, we fit the V-value to the current best to consider the influence of the Q-value. Then, we consider in-distribution and OOD action for the Q-value of different state-action. The penalty term penalizes the overestimation of OOD action to make it as small as possible. For the case where all state-action pairs are in-distribution, we take the derivative with respect to Equation \ref{eq:QL_tau}: 
\begin{equation}
    \nabla_{\theta}L_Q(\theta) = \nabla_{\theta}Q(s, a)\times(\left(r(s,  a)+\gamma V_{\psi}\left(s^{\prime}\right)-Q_{\theta}(s,  a)\right) + \beta \times \nabla_{\theta}Q(s, a)
    \label{eq:Q_Ld}
\end{equation}
Through this equation, we get the Q-value of in-distribution state-action that minimizes objective loss. As a result, for each iteration, we have the following equation:  
\begin{equation}
    Q_{k}(s, a) = r + \gamma  V_{k}(s^{\prime}) - \beta \label{eq:Q}
\end{equation}
Then, we fit the Q-value to the current best to consider the influence of the V-value. Equation \ref{eq:V_L} shows that V-function is only affected by the Q-value of in-distribution state-actions so that we can get a relatively accurate V-value estimation of state.

Given the above result $Q_k(s, a)=r+\gamma V_k(s^{\prime})-\beta$, we substitute it into the equation of V-function $V(s)=\mathbb{E}_{(s, a)\sim \mathcal{D}}(Q(s, a))$. As a result, we can get follow equation: 
\begin{equation}
    \begin{aligned}
    V_{k+1}(s) &= \mathbb{E}_{(s, a)\sim \mathcal{D}}(Q(s, a))=\mathbb{E}_{(s, a)\sim \mathcal{D}}(r+\gamma V_k(s^\prime)-\beta)=\mathbb{E}_{(s, a)\sim \mathcal{D}}(r+\gamma V^*_k(s^\prime)-\frac{1-\gamma^k}{1-\gamma}\beta)\\
    &=\mathbb{E}_{(s, a)\sim \mathcal{D}}(Q^*(s,a)-\frac{1-\gamma^k}{1-\gamma}\beta)=V^*_{k+1}(s) - \frac{1-\gamma^k}{1-\gamma}\beta \label{eq:V}
\end{aligned}
\end{equation}
$Q^*_k(s, a)$ represents Q-value estimated by the original Q-function in the previous iteration. $V^*_{k+1}(s)$ represents the current V-function's optimal estimation, corresponding to the original V-value, which is estimated by V-function trained by Equation \ref{eq:Q_L}. We can iterate through the concept of V and Equation \ref{eq:Q} to prove the conclusion of Equation \ref{eq:V}. For a static $\beta$, the final V-value is: $V(s)=V^*(s)-\frac{1-\gamma^n}{1-\gamma}\beta$, in which n means that it takes n loops of the V function to converge.

Let's plug $V(s)=V^*(s)-\frac{1-\gamma^n}{1-\gamma}\beta$ into Equation \ref{eq:Q_Ld} to consider the optimal Q-value: 
\begin{equation}
    Q(s, a)=r+\gamma V(s^{\prime})-\beta
    =Q^*(s, a)-\frac{1-\gamma^{(n+1)}}{1-\gamma}\beta \label{eq:Q_f}
\end{equation}
$V^*$ represents the final V-value estimated by the original V-function, and $Q^*$ has a similar meaning. Therefore, we get our algorithm's theoretical Q-value and V-value, and our work shows that different environments have different n.

According to the above results, we get a relatively accurate Q-value and V-value of in-dataset state-action and a conservative Q-value of OOD actions. By adjusting the parameter $\beta$, we can change the degree of OOD actions' estimation and may have a higher probability of choosing OOD action.

\textbf{Performance Analysis. }We will provide an analysis of whether our algorithm can achieve state-of-the-art performance. First, for critics with V-function and Q-function, we know that it can recover the optimal value function under the dataset support constraints, which had been analysed in IQL. Therefore, our algorithm can sample the action which is best in the dataset.

However, we want our algorithm to explore appropriately to find a better policy distribution than an in-distribution optimal policy if it exists. We assume that better policy action is near the optimal policy action within the dataset and can be sampled if it exists. Our algorithm tries to reduce the Q-value of the action sampled by the current policy.

For optimal in-distribution action, our algorithm estimate the Q-value of it as $Q(s, a)=Q^*(s, a)-\frac{1-\gamma^{(n+1)}}{1-\gamma}\beta$. We assume that there is an OOD action which has a higher Q-value than the one above, i.e. $Q^*(s, a^\prime)>Q^*(s, a)$. For this action, the theoretical Q-value of our algorithm is $Q(s, a^\prime)=Q^*(s, a^\prime)-\frac{1-\gamma^{(n_1+1)}}{1-\gamma})\beta$ (the rate of convergence of different Q may be different). As a result, we show that $Q(s, a^\prime)$ should be selected if $Q^*(s, a^\prime)>Q^*(s, a)+(\frac{1-\gamma^{(n+1)}}{1-\gamma} - \frac{1-\gamma^{(n_1+1)}}{1-\gamma})\beta$. Therefore, our algorithm has some exploratory ability to determine whether there is a better policy near the in-dataset optimal policy distribution to perform better.

In practice, as we show in sections 5.1 and 5.3, the actual Q-value floats around the theoretical Q-value. We analyse the difference bound between the actual Q-values and theoretical Q-values. At first, if most of the in-dataset actions will be not fully sample, the Q-value will be: $Q^*(s,a)>\hat{Q}(s,a)>Q^*(s,a)-\frac{1-\gamma^{(n+1)}}{1-\gamma}\beta$. In addition, we analyse the final bound in Appendix C.

\subsection{Implementation}
\begin{algorithm}[H]
    \caption{flexible conservative estimate}
    \begin{algorithmic}
        \label{alg:iql}
        \State Initialize V network $\psi$, Q network $\theta$, target Q network $\hat{\theta}$ and update target Q network $\hat{\theta} \leftarrow \theta$,  action network $\phi$, set expectile regression parameters $\tau$, $\alpha$, and conservative parameter $\beta$ according to the dataset. 
        \State TD learning (DCE): 
        \For{each epoch}
        \State use Equation \ref{eq:V_L} to update the parameter of V-function $\psi$
        \State use Equation \ref{eq:our} to update the parameter of Q-function $\theta$ and update target network parameter 
        \State $\hat{\theta} \leftarrow (1-\gamma)\hat{\theta} + \gamma\theta$
        \State If necessary, update $\alpha$ by SAC $\alpha$ loss and update $\beta$ by your update rule(in our algorithm, we use linear update every 50 epochs)
        \EndFor
        \State Policy extraction: 
        \For{each epoch}
        \State Update policy by SAC policy loss
        \EndFor
    \end{algorithmic}
\end{algorithm}
We will discuss the implementation details of this algorithm. While our algorithm uses a similar critic loss objective, we do not use the advantage-weighted regression to train the actor. Our algorithm only makes a few simple changes to the general Actor-Critic. We use the standard SAC (\cite{haarnoja2018soft}) policy as the policy of our work. I.e.the policy loss of our algorithm is as follows: 
\begin{equation}
L_\pi(s)=-\frac{1}{|\mathcal{B}|} \sum_{\left(s_{t},  a_{t},  r_{t+1},  s_{t+1}\right) \in \mathcal{B}} E_{a_{t}^{\prime} \sim \pi\left(\cdot \mid s_{t} ; \theta\right)}\left[q_{0}\left(s_{t},  a_{t}^{\prime}\right)-\alpha \ln \pi\left(a_{t}^{\prime} \mid s_{t} ; \theta\right)\right] \label{equa 12}
\end{equation}
Where $B$ is the dataset, and $\theta$ is the policy parameter. In practice, policy loss of our algorithm does not require$-\frac{1}{|\mathcal{B}|}$.

In addition, based on the IQL critique, our work introduces an additional loss to achieve flexible conservative estimation. Specific, first, we train V-function and Q-function by using Equations \ref{eq:V_L} and \ref{eq:our}. At the same time, we update the parameter $\alpha$ if the configure required (It is usually decided by dataset and the relationship between $\alpha$ and $\beta$ discussed in Appendix B). Then, we use Equation \ref{equa 12} to train the policy with the parameter $\alpha$. Although we use static parameter $\beta$ in the above subsection, we use a variable $\beta$ to get better performance. Specific, we change $\beta$ when Q-function and V-function converges, and by selecting appropriate $\beta$ for different dataset and environment, our algorithm can have state-of-the-art performance.

\section{Experiments}

\begin{wrapfigure}{R}{0.5\textwidth}
    \centering
    \includegraphics[width=0.5\textwidth]{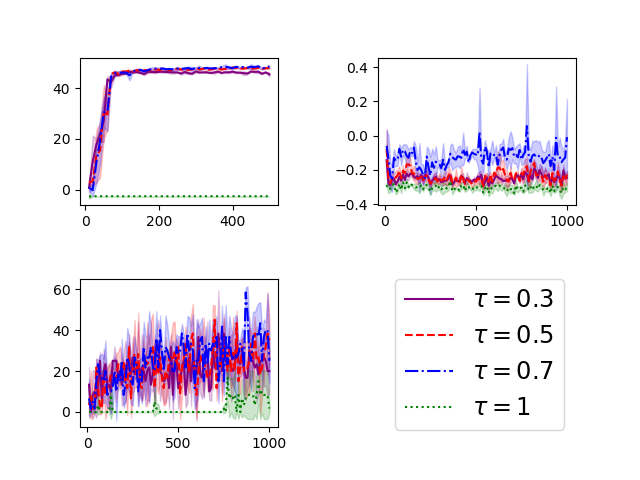}
    \caption{This figure shows the performance of different $\tau$. \textbf{top left:} The performance in halfcheetah-medium. We can see that expect some values, different $\tau$ show little difference (Appendix E try analysis this phenomenon). \textbf{top right:} The performance in relocate-cloned. \textbf{lower left:} The performance in kitchen-mixed. By this figure, we find that in the Adroit and Kitchen task, the value of $\tau$ will impact the performance of our algorithm. 
    }
    \label{fig:tau}
\end{wrapfigure}
In experiments, we demonstrate the performance of DCE in various settings and compare it to prior offline RL methods. First, we start with simple ablation experiments to show that our method has an effect and that the coefficient $\beta$ can control the degree of conservatism. In addition, we also prove the correctness of the changes in the Q-value of in-dataset state-action and V values analyzed in the previous section. Then, we will compare DCE with state-of-the-art offline RL methods on the D4RL (\cite{fu2020d4rl}) benchmark tasks.

Before experiments, we try the different parameter $\tau$ to show if the influence of $\tau$ is different in our algorithm (see Figure \ref{fig:tau}). Figure \ref{fig:tau} shows the performance of different $\tau$ in halfcheetah-medium, relocate-cloned, kitchen-mixed. The different values of $\tau$ affect the results of our algorithm. In addition, we have different degrees of influence of $\tau$ for different environments, and we try to analyze this phenomenon in Appendix E. For simplicity, our algorithm uses the same configure of $\tau$ in experiments.

\begin{figure}
\centering
\begin{subfigure}
    \centering
    \includegraphics[width=0.32\textwidth]{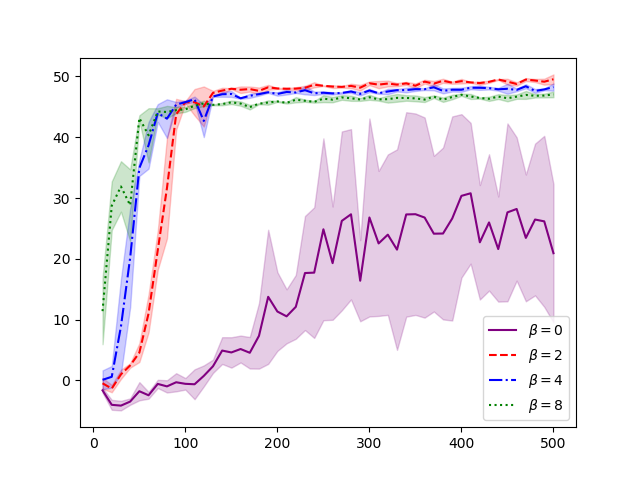}
\end{subfigure}
\hspace{0.01mm}
\begin{subfigure}
    \centering
    \includegraphics[width=0.32\textwidth]{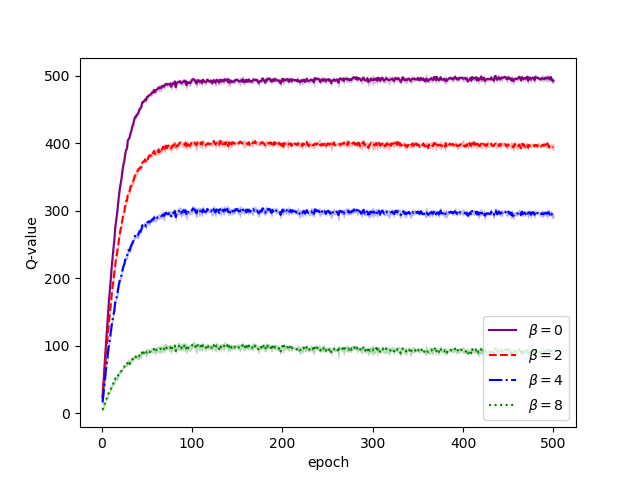}
\end{subfigure}
\hspace{0.01mm}
\begin{subfigure}
    \centering
    \includegraphics[width=0.32\textwidth]{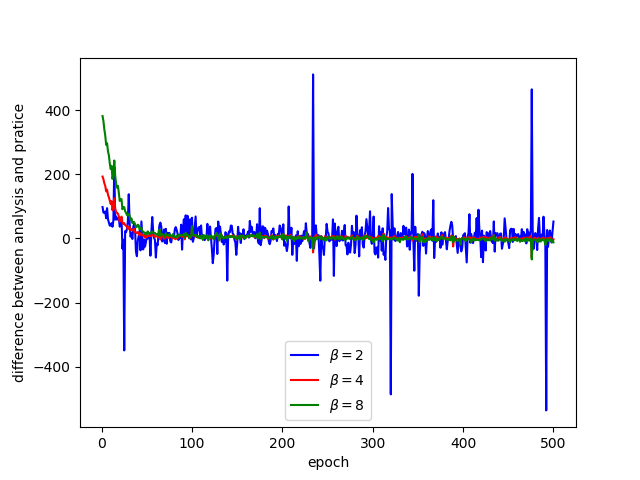}
\end{subfigure}
    \caption{Our algorithm simply runs 500 epochs on a given random factor as an ablation experiment. \textbf{left: }We show the average normalized return of our algorithm with different parameters $\beta$ in this sub-graph. \textbf{middle: }We show the average Q-value of in-dataset state-action for different $\beta$ in this sub-graph. It is obvious that Q-value varies with the parameter $\beta$. \textbf{right: }We computer the difference between Q-value of in-dataset state-action and $r+\gamma V^*-\frac{1-\gamma^{(n+1)}}{1-\gamma}\beta$ and show the results on this sub-graph. This result shows that the Q-value of in-dataset state-action obeys our analysis in the previous section. }
    \label{fig:in-dis ac q}
\end{figure}
\subsection{Double Conservative Estimation Of In-dataset Action}
To verify that our penalty term is valid,   we control for additional parameters to experiment. In this experiment,  we use different $\beta$ and the same seed as a compare condition. For $\alpha$ in SAC,  we fixed this parameter and get $\alpha=1.0$. To show the effect of our method,  we set $\beta = 0,  2,  4,  8$ and show the performance of different $\beta$. For other parameters like t in V learning, we use the same sets with IQL. We run this network for 500000 gradients in halfcheetah-medium-v2 datasets, and the result is drawn as shown in Fig.\ref{fig:in-dis ac q}. 

We can see that $\beta=0$, this method has a poor performance even is lower than BC, and an inappropriate value of $\beta$, for example, 8, will have a poor performance too. However, even with the parameter $\beta=0$, the policy performance is better than traditional SAC, which means the introduction of the V-function has a certain effect on preventing policy extrapolation. In addition, it is obvious that the Q values of in-dataset state-action change proportionally follow $\beta$ (seeing the middle of Fig.\ref{fig:in-dis ac q}). This means,  as shown in the right-hand sub-graph in Fig.\ref{fig:in-dis ac q}, that the Q-value of in-distribution state-action varies with the value of parameter $\beta$ and that the magnitude of the change conforms to our analysis above (i.e. the Q-value of in-dataset state-action fluctuates around $r+\gamma V(s)-\frac{1-\gamma^{(n+1)}}{1-\gamma}\beta$). At the same time, as shown in the figure, there is a big wave in the in-distribution Q-value for some value of $\beta$ ($\beta=2$). And from that, we can deduce that the relationship between value of V and $V^*$ also satisfies the above analysis because of $V(s) = \mathbb{E}_{(s, a)\sim \mathcal{D}}(Q(s, a))$. Of course, the performance of policy change is negligible when the $\beta$ has a large value.

\subsection{Offline RL Benchmarks}
In this subsection, we compare our algorithm with other offline RL algorithms in the D4RL benchmark (see Table \ref{table:total results}). In D4RL, there are four types of MuJoCo tasks: Gym locomotion tasks, the Ant Maze tasks, the Adroit tasks and Kitchen robotic manipulation tasks. Our work compares the baseline algorithms on the Gym and Adroit environments, which have been used extensively in previous research. We compare our DCE with a series of representative algorithms: BC, UWAC (\cite{wu2021uncertainty}), PBRL (\cite{bai2022pessimistic}), IQL (\cite{kostrikov2021offline}), TD3-BC (\cite{fujimoto2021minimalist}), CQL (\cite{kumar2020conservative}). We analyse the performance of the algorithm on Gym locomotion tasks in this section and the algorithm's performance on Adroit tasks in Appendix A.

\textbf{Result in Gym locomotion tasks. }The Gym locomotion tasks include three environments: HalfCheetah,  Hopper and Walker2D, and five dataset types for every environment: random,  medium,  medium-replay,  medium-expert and expert. Our experiments run our algorithm on the first four datasets with the 'v2' type. We train each algorithm for one million time steps and record the average policy performance by interacting with the online environment. Table \ref{table:total results} reports the average normalized score corresponding to the performance of the different algorithms in each task. This table shows that PBRL performs best in most tasks among all the baseline algorithms, and CQL, IQL, and TD3-BC perform competitively in medium-expert datasets. Furthermore, we found that random seeds influence our algorithm in some datasets.

Through careful observation, we found that compared with baseline algorithms,  our approach performs best on each dataset of the HalfCheetah, and the effect of random seed is small. In addition, compared with baseline algorithms, our algorithm usually has a better performance in the datasets without any optimal solution, including random, medium, and medium replay. In addition, in the medium-expert dataset of each task, our algorithm has no significant advantage over PBRL, IQL, TD3-BC and CQL.

\begin{table}[t]
\centering
\scriptsize
\begin{tabular}{l||rrrrrr|r}
\toprule
Dataset &BC    & UWAC   & PBRL   &IQL & TD3-BC &CQL & \textbf{Ours}\\\hline
\midrule
HalfCheetah-r  &2.1  &2.3 $\pm$0.0  &11.0 $\pm$5.8  &13.7$\pm5.3$  &11.0 $\pm$5.8  &17.5 $\pm$1.5 &\textbf{23.5$\pm$3.3} \\\hline
    
Hopper-r       &9.8 &2.7$\pm$0.3 &\textbf{26.8$\pm$9.3} &7.1$\pm1.7$ &8.5$\pm$0.6 &7.9$\pm$0.4 &9.6$\pm$3\\\hline

Walker2d-r     &1.6 &2.0$\pm$0.4 &8.1$\pm$4.4 &7.6$\pm0.9$ &1.6$\pm$1.7 &5.1$\pm$1.3 &\textbf{11.3$\pm$6.6} \\\hline
\midrule
HalfCheetah-m  & 42.6  & 42.2 $\pm$0.4  &\textbf{57.9$\pm$1.5}  &47.4$\pm$0.2  &48.3 $\pm$0.3  &44.0$\pm$5.4 &\textbf{57.4$\pm$1.3} \\\hline
    
Hopper-m       &52.9 &50.9$\pm$4.4 &\textbf{75.3$\pm$31.2} &66.2$\pm$5.7 &59.3$\pm$4.2 &58.5$\pm$2.1 &\textbf{75.9$\pm$12.3}\\\hline

Walker2d-m     &75.3 &75.4$\pm$3.0 &\textbf{89.6$\pm$0.7} &78.3$\pm$8.7 &83.7$\pm$2.1 &72.5$\pm$0.8 &84.7$\pm$4.5 \\\hline

\midrule
HalfCheetah-m-r &36.6 &35.9$\pm$3.7 &45.1$\pm$8.0 &44.2$\pm$1.2 &44.6$\pm$0.5 &45.5$\pm$0.5 &\textbf{53.4$\pm$1.3}\\\hline

Hopper-m-r  &18.1 &25.3$\pm$1.7 &\textbf{100.6$\pm$1.0} &94.7$\pm$8.6 &60.9$\pm$18.8 &95.0$\pm$6.4 &\textbf{99.5$\pm$8.6} \\\hline

Walker2d-m-r  &26.0 &23.6$\pm$6.9 &77.7$\pm$14.5 &73.8$\pm$7.1 &\textbf{81.8$\pm$5.5} &77.2$\pm$5.5 &\textbf{82.3$\pm$11.7} \\\hline

\midrule
HalfCheetah-m-e &55.2 &42.7$\pm$0.3 &\textbf{92.3$\pm$1.1} &86.7$\pm$5.3 &90.7$\pm$4.3 &\textbf{91.6$\pm$2.8} &\textbf{92.7$\pm$1.6}  \\\hline

Hopper-m-e   &52.5 &44.9$\pm$8.1 &\textbf{110.8$\pm$0.8} &91.5$\pm$14.3 &98.0$\pm$9.4 &105.4$\pm$6.8 &91.3$\pm$18.4\\\hline
   
Walker2d-m-e  &107.5 &96.5$\pm$9.1 &\textbf{110.1$\pm$0.3} &\textbf{109.6$\pm$1.0} &\textbf{110.1$\pm$0.5} &108.8$\pm$0.7 &\textbf{110.2$\pm$0.7}\\\hline
\midrule

\textbf{MuJoCo total} &480 &444.4$\pm$38.3 &\textbf{805.3$\pm$78.6} &720.8$\pm$60 &698.5$\pm$53.7 &729$\pm$34.2 &\textbf{791.7$\pm$73.3}\\\hline
\bottomrule
\textbf{Adroit total} &93.9 &34.8$\pm$11 &116.1$\pm9.4$18.7 &118.1$\pm$30.7 &0.0 &93.6 &\textbf{159.3$\pm$85.1}\\\hline
\textbf{Kitchen total} &\textbf{154.5} &- &- &\textbf{159.8$\pm$22.6} &- &144.6 &\textbf{156.8$\pm$32.8}\\\hline
\bottomrule
\textbf{runtime(s)} &6 &176 &- &10 &- &18 &12
\end{tabular}
\vspace{-1em}
\caption{Average normalized score and the standard deviation of all algorithms over five seeds in Gym. For every seed, we sample four trajectories and calculate the average return of these trajectories during the evaluation period. The highest performing scores are highlighted. The score of UWAC, TD3-BC, PBRL, CQL, BC (except for the random environment), IQL (except for the random environment) is the reported scores in Table 1 of PBRL (\citet{bai2022pessimistic}) and Table 1 of IQL (\cite{kostrikov2021offline}).  The scores for other baselines are obtained by re-training with the `v2' dataset of D4RL.  Runtime shows each algorithm's computation time running 1000 iterations on RTX3070. } \label{table:total results}
\vspace{-1em}
\end{table}

\begin{figure}
    \centering
\begin{subfigure}
    \centering
    \includegraphics[width=0.32\textwidth]{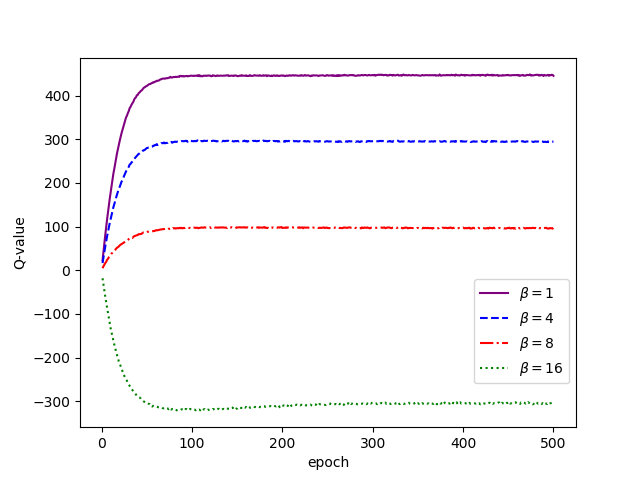}
\end{subfigure}
\hspace{0.01mm}
\begin{subfigure}
    \centering
    \includegraphics[width=0.32\textwidth]{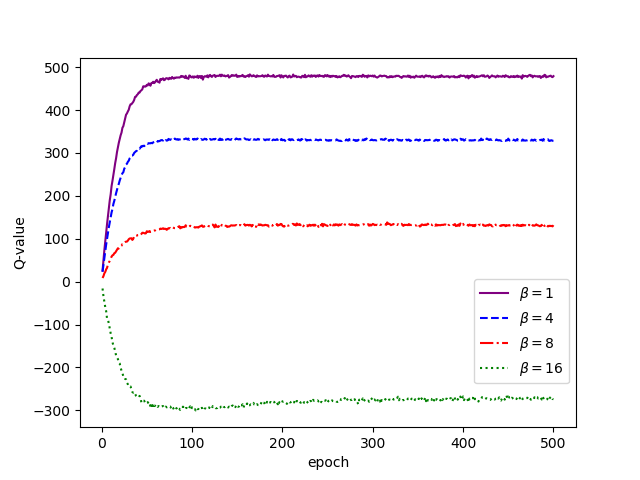}
\end{subfigure}
\hspace{0.01mm}
\begin{subfigure}
    \centering
    \includegraphics[width=0.32\textwidth]{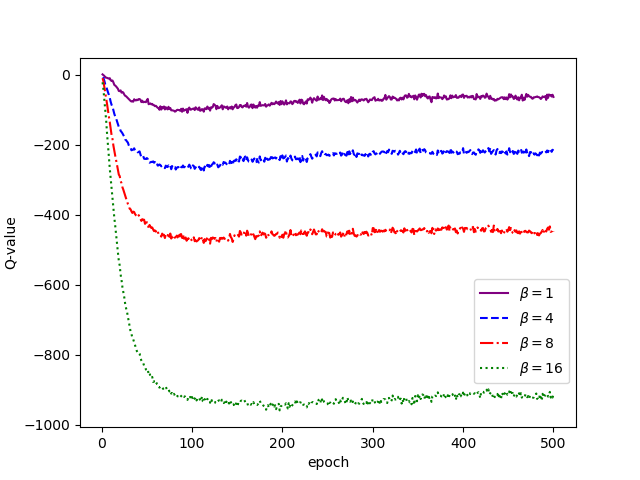}
\end{subfigure}
    \caption{We train a traditional SAC (with V-function) algorithm without any additional penalty and add four critics with different values of $\beta$. We plotted the difference between the values of traditional SAC critics and the values of critics with different values of $\beta$. \textbf{left: }We draw the Q-values of action, which is a sample by policy for different parameters $\beta$. \textbf{middle: }The difference between max Q-values of the OOD action and Q-value of in-dataset action. \textbf{right: }We show the difference between the min Q-values of the OOD action and the Q-value of the in-distribution action. }
    \label{fig:ood q}
\end{figure}

\subsection{Double Conservative Estimation Of OOD Action}
We also compare the Q-value of OOD action for different parameters $\beta$ to show that our work is also valid for OOD action and can adjust the degree of conservative estimation. We set $\beta=1, 2, 4, 8$. In order to analyze the conservatism of Q-function to OOD state-action, we add two additional policies used to provide the OOD action. In addition, we use deterministic policy to ensure that the policy provides the same action for different parameters $beta$.

Specifically, we train our algorithm while two additional policies do not require any training. One of their policies is the pre-trained best policy, and the other is the initialization policy. Their policies give a deterministic action for a state, and critics compute the corresponding Q-value. We compute the difference between the Q-value of OOD state-action generated by different $\beta$ values and the Q-value of traditional SAC critic. The result is drawn as shown in Fig.\ref{fig:ood q}. For different parameters $\beta$, OOD state-action. We find that our algorithm can control the degree of conservative estimation by adjusting parameter $\beta$. 

\subsection{Runtime}
Our algorithm has a relatively good computational speed while ensuring competitive performance. We measure runtime for our algorithm and CQL and IQL, which implementation in PyTorch both. By running those algorithms on NVIDIA GeForce RTX 3070, we found that our algorithm has faster computationally than CQL and competitive computationally for IQL. Specific, the CQL takes 18 seconds to perform 1000 gradient updates, and the IQL takes 10 seconds, while our algorithm needs 12 seconds (see Table \ref{table:total results}). 

\section{Conclusion}
In this paper, we propose a double conservative estimation, a conservative-based Model-free algorithm. This algorithm can change the degree of conservative estimation by controlling the parameter $\beta$ to get better performance when the agent meets a different environment. In addition, this algorithm avoids the influence of OOD action on in-distribution action as much as possible while reducing the value of OOD state-action. The introduction of V-function conservative estimate the Q-value of all state-action. Therefore we can see the V-function as a conservative estimation. We give a theoretical analysis of the influence of parameters on Q-function and V-function and experiments that demonstrates the state-of-the-art performance by changing the degree of conservative estimation. In the experiment, parameter $\alpha$ of control policy exploration is influenced by the parameter $\beta$ (Appendix B). However, our work lacks further analysis of the relationship between $\alpha$ and $\beta$. Therefore, our future research may focus on the relationship between the degree of conservative and policy exploration. In addition, trying to tune parameter $\beta$ automatically is also a viable direction.

\newpage
\bibliography{iclr2023_conference}
\bibliographystyle{iclr2023_conference}

\newpage
\appendix
\section{Experiments In Adroit And Kitchen Domain}
The Adroit tasks include the four environments: Pen, Hammer, Door, Relocate and three dataset types for every environment: human, clone and expert. Our experiments use two of those three datasets types: human and clone. The Kitchen tasks include complete, partial and mixed datasets. For every dataset, we use five different random seeds and sample five evaluation trajectories to average the mean performance of our algorithm. Our algorithm uses the standard offline reinforcement learning parameter and network structure. We use the same setting as IQL for parameter $\tau$. Instead of IQL, our algorithm adds dropout to the Q-function in some datasets than to policy in all datasets for Adroit tasks to get better performance.

\begin{table}[t]
\centering
\begin{tabular}{l||rrrrrr|r}
\toprule
Dataset &BC    & UWAC   & PBRL   &IQL & TD3-BC &CQL & \textbf{Ours}\\\hline
\midrule
Pen-human  &34.4  &10.1 $\pm$3.2  &35.4 $\pm$3.3  &71.5  &0.0   &37.5  &\textbf{85.1$\pm$37} \\\hline
    
Hammer-human       &1.5 &1.2$\pm$0.7 &0.4$\pm$0.3 &1.4 &0.0 &\textbf{4.4} &2.1$\pm$2.0\\\hline

Door-human     &0.5 &0.4$\pm$0.2 &0.1 &4.3 &0.0 &\textbf{9.9} &4.8$\pm$14 \\\hline
Relocate-human       &0.0 &0.0 &0.0 &0.1 &0.0 &\textbf{0.2} &0.1$\pm$0.1\\\hline
\midrule
Pen-cloned  & 56.9  & 23.0 $\pm$6.9  &\textbf{74.9$\pm$9.8}  &37.3  &0.0  &39.2 &63.5$\pm$25.7\\\hline
    
Hammer-cloned       &0.8 &0.4$\pm$0.0 &0.8$\pm$0.5 &\textbf{2.1} &0.0 &\textbf{2.1} &1.0$\pm$0.6\\\hline

Door-cloned     &-0.1 &0.0 &\textbf{4.6$\pm$4.8} &1.6 &0.0 &0.4 &2.9$\pm$5.6 \\\hline
    
Relocate-cloned       &-0.1 &-0.3 &-0.1 &-0.2 &\textbf{0.0} &-0.1 &-0.2$\pm$0.1\\\hline

\midrule

\textbf{adroit total} &93.9 &34.8$\pm$11 &116.1$\pm$18.7 &118.1 &0.0 &93.6 &\textbf{159.3$\pm$85.1}\\\hline

\midrule
kitchen-complete &\textbf{65.0} &- &- &\textbf{62.5} &- &43.8 &50$\pm$10\\\hline
kitchen-partial &38.0 &- &- &46.3 &- &49.8 &\textbf{59 $\pm$14}\\\hline
kitchen-mixed &\textbf{51.5} &- &- &\textbf{51.0} &- &\textbf{51.0} &48.8$\pm8.8$\\\hline

\textbf{kitchen total} &\textbf{154.5} &- &- &\textbf{159.8} &- &144.6 &\textbf{156.8$\pm$32.8}\\\hline

\bottomrule
\end{tabular}
\vspace{-1em}
\caption{This table shows the average normalized score and the standard deviation of all algorithms over five seeds in the Gym. For every seed, we sample four trajectories and calculate the average return of these trajectories during the evaluation period. The highest performing scores are highlighted. The score of UWAC, TD3-BC, PBRL, CQL, BC, IQL is the reported scores in Table 4 of PBRL (\citet{bai2022pessimistic}) and Table 3 of IQL (\cite{kostrikov2021offline}). In addition, the total adroit average normalized score of IQL is $118.1 \pm 30.7$, and the total kitchen average normalized score of IQL is $159.8 \pm 22.6$ in their paper, but we did not find the exact value of each task.} \label{table:AandK results}
\vspace{-1em}
\end{table}

Table \ref{table:AandK results} shows the performance of our algorithm and baseline offline reinforcement learning. Our algorithm has a better performance compared to the baseline algorithm. However, in some environments, our algorithm is sensitive to random seed (i.e. having a higher variance) because our algorithm has a constant value of parameter $\alpha$. Setting a constant value of $alpha$ is when we use auto-update $\alpha$ (see Appendix B).

\section{Relationship Between $\alpha$ And $\beta$}

\begin{figure}
    \centering
\begin{subfigure}
    \centering
    \includegraphics[width=0.4\textwidth]{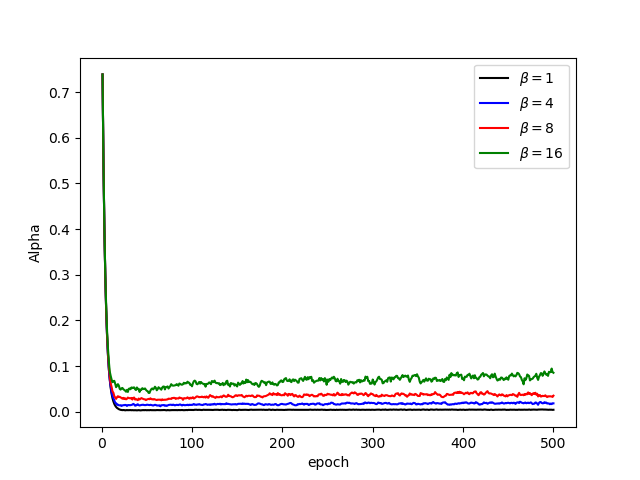}
\end{subfigure}
\hspace{0.01mm}
\begin{subfigure}
    \centering
    \includegraphics[width=0.4\textwidth]{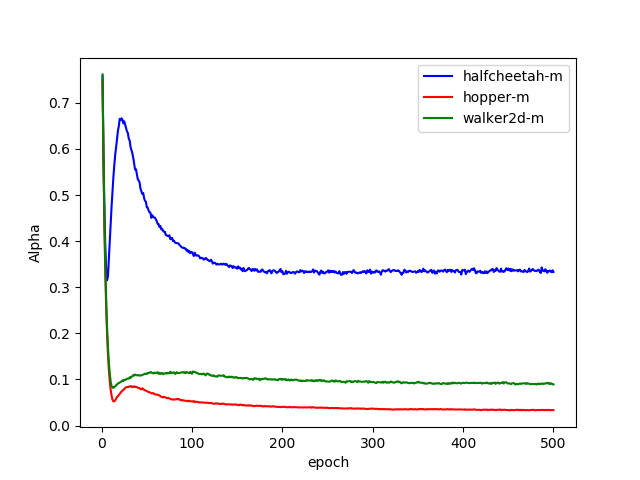}
\end{subfigure}
\hspace{0.01mm}
\begin{subfigure}
    \centering
    \includegraphics[width=0.4\textwidth]{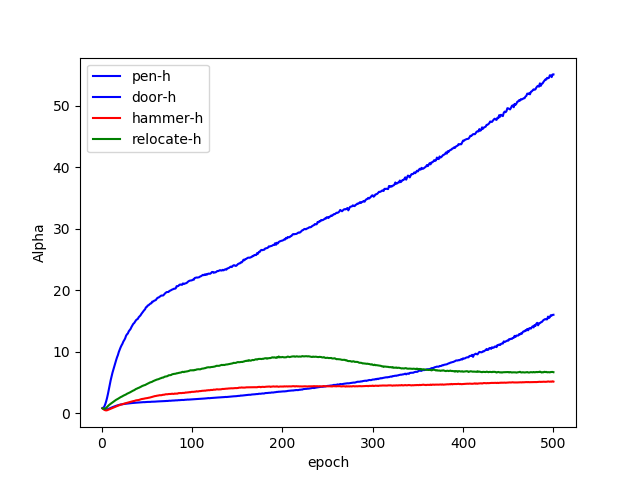}
\end{subfigure}
\begin{subfigure}
    \centering
    \includegraphics[width=0.4\textwidth]{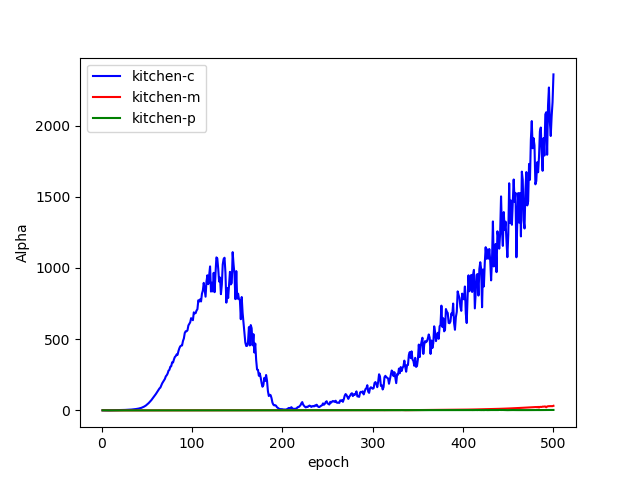}
\end{subfigure}
    \caption{We show the value of auto-update $\alpha$ for different values of $\beta$ and dataset. \textbf{top left:}We draw the value of $\alpha$ for different $\beta$. \textbf{top right:}This sub-figure show the value of $\alpha$ in the MuJoCo environment. \textbf{lower left:}The value of $\alpha$ in the Adroit environment have some differences for the different task. \textbf{lower right:}In Kitchen environment,different datasets has vastly different $\alpha$. }
    \label{fig:alpha}
\end{figure}

Unlike the IQL algorithm, which uses the advantage-weighted regression as a policy, our algorithm uses the SAC algorithm's policy. Therefore, our algorithm needs to consider two parameters $\alpha$ and $\beta$, to get a better performance in a different environment and dataset. We add the $\alpha$ into a network to auto-update it in some datasets. In other datasets, we set the $\alpha$ as a constant value because of the relationship between $\alpha$, $\beta$ and the dataset.

In this section, we use auto-update $\alpha$ in the different datasets to show the influence of parameter $\beta$ on the parameter $\alpha$. We train our algorithm with different parameters $\beta$ for 500 epochs. In addition, we train our algorithm with the same parameter $\beta$ in different environments and datasets because it has some influence on the parameter $\alpha$.

In fig.\ref{fig:alpha}, we demonstrate that the parameter $\alpha$ increases with the increase of parameter $\beta$ within a specific range. In addition, for different environments and the same value of $\beta$, the value of $\alpha$ and the above range differ. In the right of fig.\ref{fig:alpha}, we can find that different environments will significantly influence the value of $\alpha$.

\section{Bound Of Actual Q-value and Theoretical Q-value}
We will analyse the difference between the Actual Q-value and Theoretical Q-value. Furthermore, before that, we show the sample-based version of Equation \ref{eq:our}:
\begin{equation}
\begin{aligned}
\hat{Q}^{k+1} \leftarrow \arg \min _{Q} \alpha \cdot\left(\sum_{\mathbf{s} \in \mathcal{D}} \mathbb{E}_{\mathbf{a} \sim \mu(\mathbf{a} \mid \mathbf{s})}[Q(\mathbf{s}, \mathbf{a})]\right)
+\frac{1}{2|\mathcal{D}|} \sum_{\mathbf{s}, \mathbf{a}, \mathbf{s}^{\prime} \in \mathcal{D}}\left[\left(Q(\mathbf{s}, \mathbf{a})-\hat{\mathcal{B}}^{\pi} \hat{Q}^{k}(\mathbf{s}, \mathbf{a})\right)^{2}\right]
\end{aligned}
\end{equation}
Where $\hat{\mathbb{B}^\pi}$ denotes the Bellman operator computed using dataset D:
\begin{equation}
\forall \mathbf{s}, \mathbf{a} \in \mathcal{D}, \quad\left(\hat{\mathcal{B}}^{\pi} \hat{Q}^{k}\right)(\mathbf{s}, \mathbf{a})=r+\gamma \sum_{\mathbf{s}^{\prime}} \hat{T}\left(\mathbf{s}^{\prime} \mid \mathbf{s}, \mathbf{a}\right) \hat{V}^k(s)\label{equa 14}
\end{equation}
Furthermore, our algorithm uses expectile regression to learn V-value, which can give different weights for different Q-values with the parameter $\tau$.

We consider the bound of difference between Q and $Q^*$ in practical at first. By Equation \ref{eq:QL_tau} and Equation \ref{eq:our}, we know that the difference between Q and $Q^*$ is an additional penalty term. Therefore, the floating bound of they is influenced by the sampling error of the penalty term. The error of penalty term depends on the sampling frequency of policy and dataset, so we have:
\begin{equation}
    \left |\hat{Q}(s,a)-\hat{Q}^*(s,a) \right | \leq \frac{n_\mu}{n_\pi} \times \frac{1-\gamma^{(n+1)}}{1-\gamma} \times\beta
\end{equation}

Then we consider the low bound of actual Q-value and theoretical Q-value.  Following prior work \cite{auer2008near} \cite{osband2017posterior} \cite{kumar2020conservative}, We know two assumption as following:
\begin{equation}
|r-r(\mathbf{s}, \mathbf{a})| \leq \frac{C_{r, \delta}}{\sqrt{|\mathcal{D}(\mathbf{s}, \mathbf{a})|}}, \quad\left\|\hat{T}\left(\mathbf{s}^{\prime} \mid \mathbf{s}, \mathbf{a}\right)-T\left(\mathbf{s}^{\prime} \mid \mathbf{s}, \mathbf{a}\right)\right\|_{1} \leq \frac{C_{T, \delta}}{\sqrt{|\mathcal{D}(\mathbf{s}, \mathbf{a})|}} \label{equa 16}
\end{equation}
Because our algorithm use expectile regression to learn V-value, we know the bound of V-value is $V(s)\leq\frac{2\tau R_{max}}{1-\gamma}$. By Equation \ref{equa 14} and \ref{equa 16}, the difference between  $\mathcal{B}^\pi\hat{Q}^k$ and $\hat{\mathcal{B}}^\pi\hat{Q}^k$ can be bounded:
\begin{equation}
\begin{aligned}
\left|\left(\hat{\mathcal{B}}^{\pi} \hat{Q}^{k}\right)-\left(\mathcal{B}^{\pi} \hat{Q}^{k}\right)\right| &=\left|(r-r(\mathbf{s}, \mathbf{a}))+\gamma \sum_{\mathbf{s}^{\prime}}\left(\hat{T}\left(\mathbf{s}^{\prime} \mid \mathbf{s}, \mathbf{a}\right)-T\left(\mathbf{s} \mathbf{s}^{\prime} \mid \mathbf{s}, \mathbf{a}\right)\right) V(s^\prime)\right| \\
& \leq|r-r(\mathbf{s}, \mathbf{a})|+\gamma\left|\sum_{\mathbf{s}^{\prime}}\left(\hat{T}\left(\mathbf{s}^{\prime} \mid \mathbf{s}, \mathbf{a}\right)-T\left(\mathbf{s}^{\prime} \mid \mathbf{s}, \mathbf{a}\right)\right) V(s^\prime)\right| \\
& \leq \frac{C_{r, \delta}+\gamma C_{T, \delta} 2 R_{\max }\tau /(1-\gamma)}{\sqrt{|\mathcal{D}(\mathbf{s}, \mathbf{a})|}}
\end{aligned}
\end{equation}

At last, we get the bound of the actual Q-value as follows:
\begin{equation}
    \hat{Q}(s,a) \leq Q(s,a) + \frac{(1-\gamma^{(n+1)})n_\mu}{(1-\gamma)n_\pi}\beta + \frac{C_{r, \delta}+\gamma C_{T, \delta} 2 R_{\max }\tau /(1-\gamma)}{\sqrt{|\mathcal{D}(\mathbf{s}, \mathbf{a})|}}
\end{equation}

\section{Relationship with CQL}
In CQL section 3.1, they propose Equation:
\begin{equation}
\hat{Q}^{k+1} \leftarrow \arg \min _Q \alpha \mathbb{E}_{\mathbf{s} \sim \mathcal{D}, \mathbf{a} \sim \mu(\mathbf{a} \mid \mathbf{s})}[Q(\mathbf{s}, \mathbf{a})]+\frac{1}{2} \mathbb{E}_{\mathbf{s}, \mathbf{a} \sim \mathcal{D}}\left[\left(Q(\mathbf{s}, \mathbf{a})-\hat{\mathcal{B}}^\pi \hat{Q}^k(\mathbf{s}, \mathbf{a})\right)^2\right] \label{equa_cql}
\end{equation}
and analysis its lower bound. Because our algorithm is slightly different from this Equation, we need to analyse the specific difference and show the experiment. In Appendix C, we get the bound of our algorithm and compare it with CQL can know the difference between CQL and our algorithm.

\begin{figure}
    \centering
\begin{subfigure}
    \centering
    \includegraphics[width=0.4\textwidth]{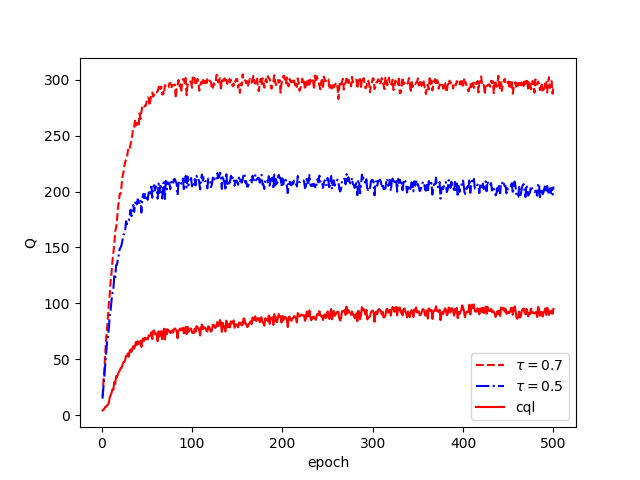}
\end{subfigure}
\begin{subfigure}
    \centering
    \includegraphics[width=0.4\textwidth]{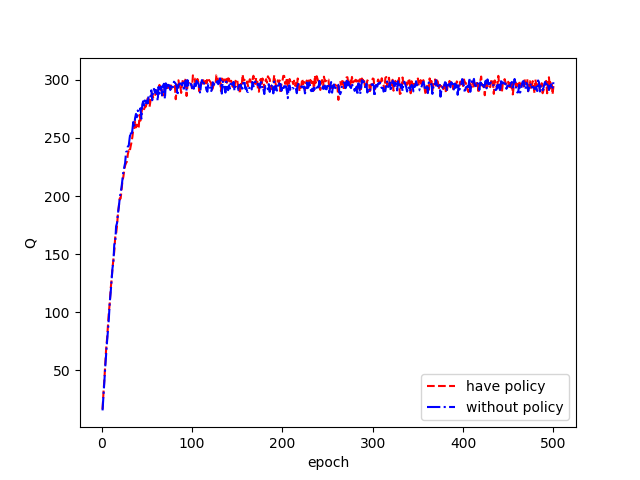}
\end{subfigure}
    \caption{There are two sub-figure to show the experiment of Appendix D. \textbf{left:}We draw the Q-value, which is obtained by different configurations of $\tau$ and Equation \ref{equa_cql} and show the difference with CQL. \textbf{right:} We show the influence of ODD action. Specific, we train critics without policy, and the penalty term is the Q-value of the current state-action.}
    \label{fig:cql Q}
\end{figure}
We run Equation \ref{equa_cql} by modifying the CQL, which sets the $\alpha$ of SAC as one and $\alpha$ of CQL as four. In addition, we use the same method as our algorithm for the penalty term rather than CQL. We compare this algorithm's Q-value with our algorithm's Q-value in Halfcheetah-medium-v2 (see Figure \ref{fig:cql Q}). We use our algorithm's different $\tau$ configured to show that parameter $\tau$ impacts the Q-value. As discussed in section 4.1, when the $\tau=0.5$, expectile regression equals standard mean squared error loss. We find that a higher $\tau$ will lead to a slack conservative estimation. In addition, we run our algorithm without a policy in which current Q-values provide the penalty term. We avoid the sampling error of the penalty term in this way. We can find that Q-value without policy will be flatter because it reduces the sampling error.

\begin{figure}
\centering
\begin{subfigure}
    \centering
    \includegraphics[width=0.32\textwidth]{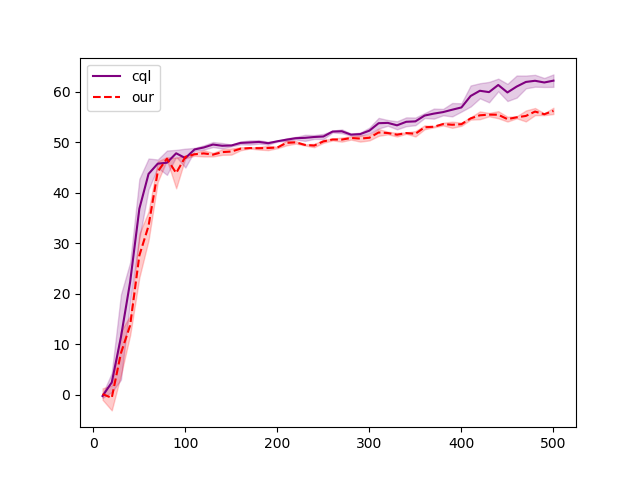}
\end{subfigure}
\hspace{0.01mm}
\begin{subfigure}
    \centering
    \includegraphics[width=0.32\textwidth]{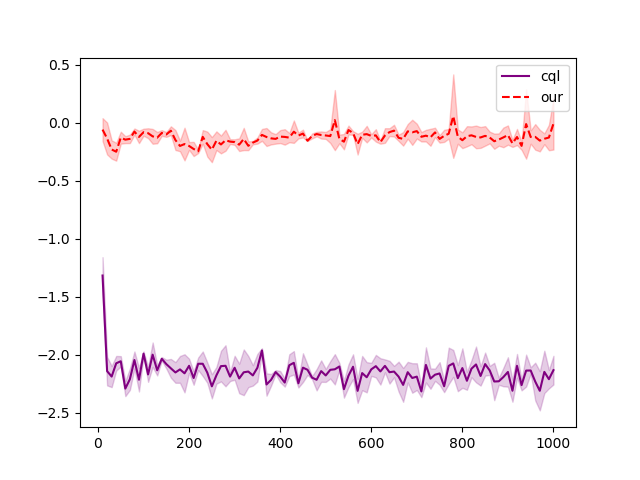}
\end{subfigure}
\hspace{0.01mm}
\begin{subfigure}
    \centering
    \includegraphics[width=0.32\textwidth]{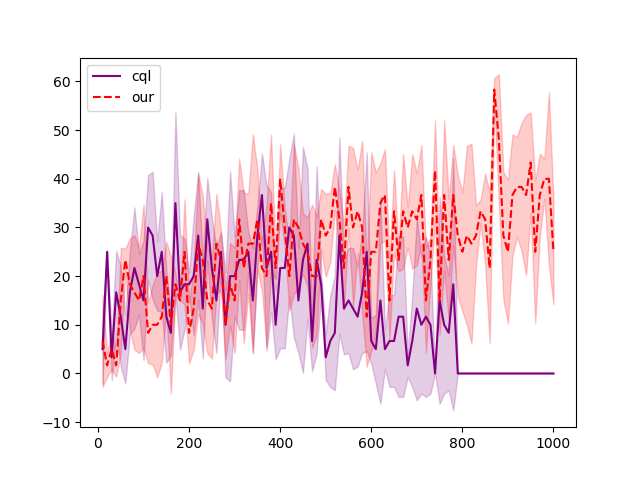}
\end{subfigure}
    \caption{\textbf{left:} The performance in halfcheetah-medium. \textbf{meidum:} The performance in relocate-cloned. \textbf{right:} The performance in kitchen-mixed.}
\label{fig:cql ret}
\end{figure}

Expect the Q-value of different Equations, and we draw the performance of their policy in different environments (see Figure \ref{fig:cql ret}). As Figure shows, in MuJoCo tasks, Equation \ref{equa_cql} of CQL is better than our algorithm, but our algorithm performs better in Adroit and Kitchen tasks. We analyse this phenomenon in Appendix E.

\section{V-function In Critic}
\begin{figure}
    \centering
\begin{subfigure}
    \centering
    \includegraphics[width=0.4\textwidth]{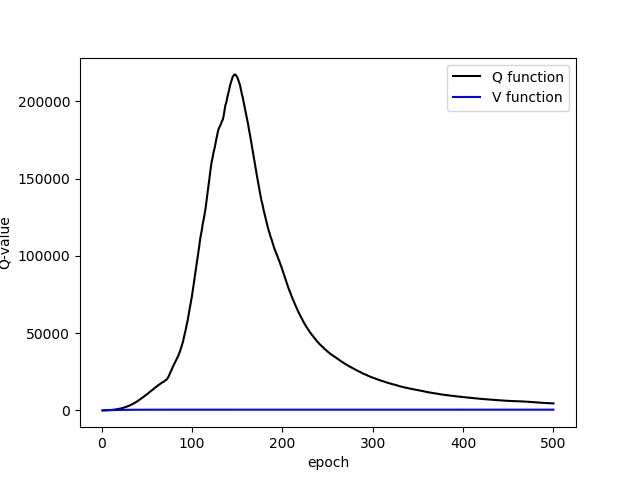}
\end{subfigure}
\begin{subfigure}
    \centering
    \includegraphics[width=0.4\textwidth]{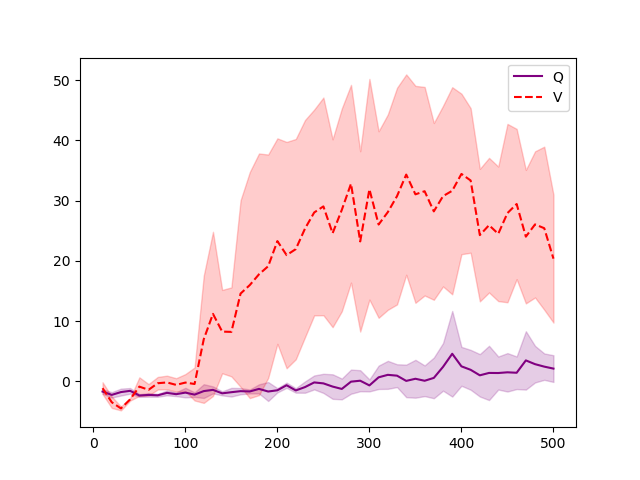}
\end{subfigure}
    \caption{\textbf{left:} Q-value of in-distribution action in halfcheetah-medium. The Q-value of SAC without V-function is 4000, and the Q-value of SAC with V-function is 400. \textbf{right:} The performance for different SAC.}
    \label{fig:SAC}
\end{figure}

\begin{wrapfigure}{R}{0.5\textwidth}
    \centering
    \includegraphics[width=0.5\textwidth]{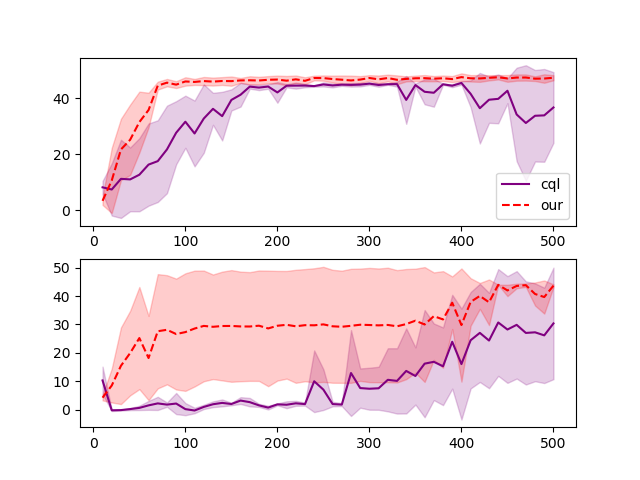}
    \caption{\textbf{left:} Average normalized return of $\alpha=3$ \textbf{right:} Average normalized return of $\alpha=5$.
    }
    \label{fig:alpha ret}
\end{wrapfigure}
In section 5, we find that parameter $\tau$ has varying degrees of influence on the performance of our algorithm. In Appendix D, the experiment shows that the agent without V-function performs better in MuJoCo tasks but worse in other tasks. We think it is caused by the V-function and the features of MuJoCo tasks.

First, V-function gives a lower value of the next state, leading to the conservative estimation of all actions. This conservative estimation limits policy exploration to a certain extent. To show this result, we train two SACs in halfcheetah-medium, one of which has a V-function (see Figure \ref{fig:SAC}).

Then, we think MuJoCo tasks have a wide `safe range' in which OOD action will have less impact than other tasks. Therefore, a looser conservative estimate would perform better. We train our algorithm and Equation \ref{equa_cql} of CQL with $\alpha=3$, and we find that although performance declined, the performance of our algorithm is better). In addition, when $\alpha=5$, Equation \ref{equa_cql} of CQL will be invalid (see Figure \ref{fig:alpha ret}). This is the exact opposite of a small $\alpha$.

\section{Configure Of Our algorithm}
We give the parameters of our algorithm under different datasets in Table \ref{table:config}. The main parameters we modify are $\alpha$ and $\beta$, so for other parameter, we use the default value (see Table \ref{table:com conifg}). In addition, we use the dropout for the critic in 'human' of Adroit tasks (value is the same as IQL). 

\begin{table}[t]
\centering
\begin{tabular}{l||rrr}
\toprule
value &alpha    &beta   &update rate \\\hline
\midrule
HalfCheetah-r  &auto  &4-0.1  &0.5\\\hline
    
Hopper-r       &auto &1-0.68 &0.016\\\hline

Walker2d-r     &auto  &0.1-0.08  &0.009 \\\hline
\midrule
HalfCheetah-m  &auto  &4-0.1  &0.5 \\\hline
    
Hopper-m       &auto  &4-0.1  &0.5\\\hline

Walker2d-m     &auto  &4-3  &0.05 \\\hline

\midrule
HalfCheetah-m-r &auto  &4-0.1  &0.5\\\hline

Hopper-m-r  &auto  &4-0.1  &3.9 \\\hline

Walker2d-m-r  &auto  &4-1  &0.15 \\\hline

\midrule
HalfCheetah-m-e &auto  &4-18  &-14  \\\hline

Hopper-m-e   &auto  &6  &0\\\hline
   
Walker2d-m-e  &auto  &4-3  &0.05\\\hline
\midrule
Pen-human  &20  &30  &0 \\\hline
Hammer-human       &auto  &5  &0\\\hline
Door-human     &0.5  &4-0.5  &0.25 \\\hline
Relocate-human       &1  &3  &0\\\hline
\midrule
Pen-cloned  &20  &10  &0\\\hline
    
Hammer-cloned &4  &4  &0\\\hline

Door-cloned     &1  &2  &0 \\\hline
    
Relocate-cloned       &1  &0.1  &0.5\\\hline
\midrule
kitchen-complete &0.5  &2-1.7  &0.015\\\hline
kitchen-partial &0.5  &2-1.65  &0.0175\\\hline
kitchen-mixed &0.5  &1.5-1.15  &0.0175\\\hline

\bottomrule
\end{tabular}
\vspace{-1em}
\caption{Configure of our algorithm. Specially, we update the $\beta$ every 50 epoch} \label{table:config}
\vspace{-1em}
\end{table}

\begin{table}[t]
\centering
\begin{tabular}{l||rr}
\toprule
Hyper-parameter &Value &Description.\\\hline
Q-network &FC(256,256) &Full Connected layers with ReLU activations.\\\hline
V-network &FC(256,256) &Full Connected layers with ReLU activations.\\\hline
Actor &FC(256,256) &Full Connected layers with ReLU activations and \\& &Gaussian distribution.\\\hline
lr of O-network &3e-4 &Q-network learning rate.\\\hline
lr of V-network &3e-4 &V-network learning rate.\\\hline
lr of actor &3e-4 &Policy learning rate.\\\hline
$\tau$ &0.7 & The value of expectile regression.\\\hline
Optimizer &Adam &Optimizer.\\\hline
$\gamma$ &0.99 &Discount factor.\\\hline
$\upsilon$ &0.005 &Target network smoothing coefficient.\\\hline

\bottomrule
\end{tabular}
\vspace{-1em}
\caption{The configuration of other parameters} \label{table:com conifg}
\vspace{-1em}
\end{table}

\end{document}